\documentclass[authoryear]{elsarticle}

\usepackage{subcaption,siunitx,booktabs}
\usepackage{multirow}
\usepackage{tabularx,booktabs} 
\usepackage{scrextend}
\usepackage{tablefootnote}
\usepackage{url}

\usepackage[breaklinks]{hyperref}
\usepackage{breakurl}

\usepackage[table]{xcolor}

\journal{Elsevier}
\biboptions{authoryear}
\usepackage[authoryear]{natbib}

\usepackage{makecell}

\begin{document}

	\begin{frontmatter}
	\title{Empirical study of the modulus as activation function in computer vision applications}
	\author[UV]{Iván Vallés-Pérez}
	\author[UV]{Emilio Soria-Olivas}

	\author[UV]{Marcelino Martínez-Sober\footnote{Corresponding author.}}
	\author[UV]{Antonio J. Serrano-López}
	\author[UV]{Joan Vila-Francés}
	\author[UV]{Juan Gómez-Sanchís}
	
	\address[UV]{IDAL, Intelligent Data Analysis Laboratory, University of Valencia, Avenida de la Universitat s/n 46100 Burjassot, Valencia, Spain. 
	}
	
	\begin{abstract}
	In this work we propose a new non-monotonic activation function: the \textit{modulus}. The majority of the reported research on nonlinearities is focused on monotonic functions. We empirically demonstrate how by using the \textit{modulus} activation function on computer vision tasks the models generalize better than with other nonlinearities - up to a 15\% accuracy increase in \textit{CIFAR100} and 4\% in \textit{CIFAR10}, relative to the best of the benchmark activations tested. With the proposed activation function the vanishing gradient and dying neurons problems disappear, because the derivative of the activation function is always 1 or -1. The simplicity of the proposed function and its derivative make this solution specially suitable for \textit{TinyML} and hardware applications. 
	\end{abstract}
	
	\begin{keyword}
		Deep Learning \sep Activation Functions \sep Optimization
	\end{keyword}
	
\end{frontmatter}

\section{Introduction}
The core piece of all deep learning models is the activation function. They enable the models to produce non-linear abstract representations when applying linear transformations in cascade \citep{goodfellow2016}. The most common choice in modern deep learning models is the Rectified Linear Unit (\textit{ReLU}) \citep{nair2010}. Among other advantages, \textit{ReLU} nonlinearities allowed us to train deeper neural networks \citep{xu2015}.

Apart from the \textit{ReLU} in the last years there have appeared many alternative activation functions \citep{dubey2022}. The following studies represent some of the most popular examples: \textit{Leaky-ReLU} and \textit{PR-ReLU} \citep{xu2015}, \textit{ELU} \citep{djork2016}, \textit{Swish} \citep{ramachandran2018}, \textit{SELU} \citep{klambauer2017}, \textit{F/PFLU} \citep{zhu2020}, \textit{RSigElu} \citep{Kilicarslan2021} etc. The authors of \citep{agostinelli2014} studied how to learn adaptive piecewise linear activation functions. All the mentioned studies propose functions that have the following properties in common: nonlinearity, continuity, differentiability and low computational cost (being \textit{Swish} the most computationally expensive option). The majority of the activation functions are monotonic, \textit{Swish}, \textit{GELU} \citep{hendrycks2016},  \textit{S-ReLU} \citep{Jin2016} and \textit{Mish} \citep{misra2019mish} are some of the most popular exceptions. Despite the large pool of alternatives, \textit{ReLU} still appears as the default choice in many applications due to its simplicity and low computational cost \citep{nair2010}.

In this work, we investigate the use of the \textit{modulus} function, also known as the absolute value function, as an activation function for deep learning models. Previous research \citep{karnewar2018} has hinted at the potential of the modulus nonlinearity in the design of a specific neural network architecture, but its performance has not been systematically compared to other nonlinearities. Through empirical evaluation, we provide empirical evidence proving that the modulus activation function allows deep learning models to converge to better solutions in \textit{CIFAR10}, \textit{CIFAR100} and \textit{MNIST} datasets. Compared with other modern non-monotonic nonlinearities such as \textit{Mish}, \textit{PFLU} and \textit{FPFLU} or \textit{Swish}, the \textit{modulus} has a very low computational cost (equivalent to \textit{ReLU}). This property makes the \textit{modulus} specially useful for \textit{TinyML} and hardware applications \citep{sanchez2020} and hardware neural networks \citep{Misra2010}.

The contributions of this paper are: (1) we propose the modulus activation function, (2) the proposed function empirically shows significantly better results in 75\% of our experiments, (3) we contribute to the new trend of non-monotonic activation functions with another example showing good results in the same direction, (4) the simplicity of the modulus activation function makes it very suitable for hardware applications and \textit{TinyML}, and (5) we propose two smooth approximation to the \textit{modulus} function, one of them achieving even superior results than the original \textit{modulus}.

This paper is organized as follows. Section \ref{sec:methods} describes the proposed activation function and the ones used for benchmarking. Section \ref{sec:experiments} describes the experiments conducted and the results achieved. Finally, we summarize our contribution in section \ref{sec:conclusions}, enumerating the main conclusions.


\section{Methods} \label{sec:methods}
\subsection{Modulus activation function}
In this section we introduce our proposed activation function, the \textit{modulus}: $f(x)=|x|$. This nonlinearity is a continuous, piecewise-linear function consisting of an identity mapping for positive values of $x$, and a negative identity mapping for negative values of $x$. Its derivative (defined below), as well as the modulus function itself, are hardware-level bit-size operations. This is extremely useful for hardware implementations. Besides, the modulus function can be expressed as $f(x) = \text{sgn}(x)\cdot x$, where $ f'(x) = \text{sgn}(x)$, and hence $f(x) = f'(x)\cdot x$. In practice, this means that the derivative can be calculated as part of the forward pass and cached for the \textit{backpropagation}, reducing the total computation. This form is also specially useful in hardware implementations, given that the whole activation function and its derivative is fully represented as a bit-size operation (the sign function). In hardware, the memory and computing requirements are limited, hence having mechanisms that allow for cheap computations (1-bit in the modulus activation) and reusability/caching (part of the gradients computed during the forward pass, in our case) is crucial. 

\begin{equation}
f'(x)= \mathrm{sgn}(x) \left\{ \begin{array}{lcc}
1 &   \text{if}  & x > 0 \\
-1 &  \text{if} & x < 0 
\end{array} \right.
\end{equation}

\begin{figure}
	\centering
	\includegraphics[width=1.0\linewidth]{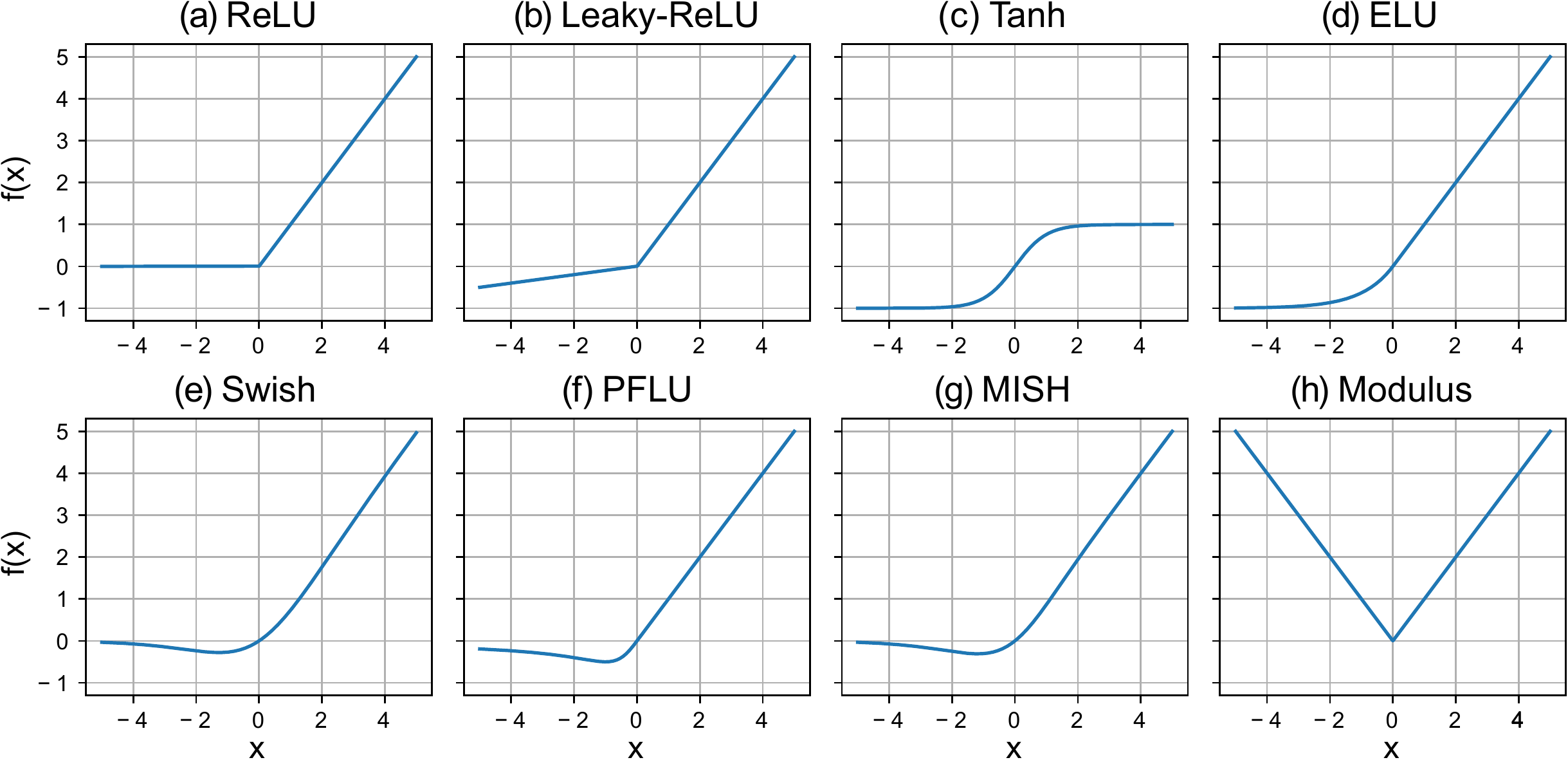}
	\caption{Nonlinearities used along this study. (a-g) are benchmark activation functions that showed good performance in previous studies. (h) is the \textit{modulus} activation function proposed in this study. For this example, $\beta=10.0$, $\beta=1.0$ and $\beta=1.0$ have been used for the \textit{Leaky-ReLU}, \textit{ELU} and \textit{Swish} activation functions, respectively.}
	\label{fig:activations}
\end{figure}

Notice that the \textit{modulus} function is differentiable everywhere except in $x = 0$. For practical purposes, we define $f'(0) = 1$ so that the derivative is defined for all the range of $x$ values, similar to the case of \textit{ReLU} \citep{goodfellow2016}, and to ensure that the norm of the gradient is constant for all values of $x$. See figure \ref{fig:activations}h for a graphical representation.

The \textit{modulus} activation function belongs, in essence, to the family of rectifier functions \citep{glorot2015} . In fact, it is equivalent to a \textit{Leaky-ReLU} with $\beta=-1$ (see equation \ref{eq:leaky-relu}). However \textit{Leaky-ReLU} was originally defined to take values of beta strictly higher than 1 \citep{xu2015}. Furthermore, we empirically show that the \textit{modulus} achieves significantly superior results than the \textit{Leaky-ReLU}.

The benefit of this activation function with respect to the other rectifiers is that the norm of its gradient is constant ($||\nabla_x f|| = 1 \quad \forall x$) and hence, interestingly, it does not depend on the value of $x$. This property is desirable when optimizing the parameters of a neural network with gradient descent algorithms, as there are no input values for which the neuron saturates or explodes \citep{glorot2010} (i.e. values of x for which the gradient of the activation function is close to zero, or extremely large). This naturally removes the dying neurons \citep{lu2020} and vanishing gradient problems \citep{pascanu13, hochreiter1998, Hochreiter2001}, which usually appears in activations with zero regions (such as \textit{ReLU}) or with asymptotically saturating regions (such as \textit{tanh}), respectively.

\subsection{Smooth approximations of the modulus function}
\begin{figure}[h!]
	\centering
	\includegraphics[width=0.65\linewidth]{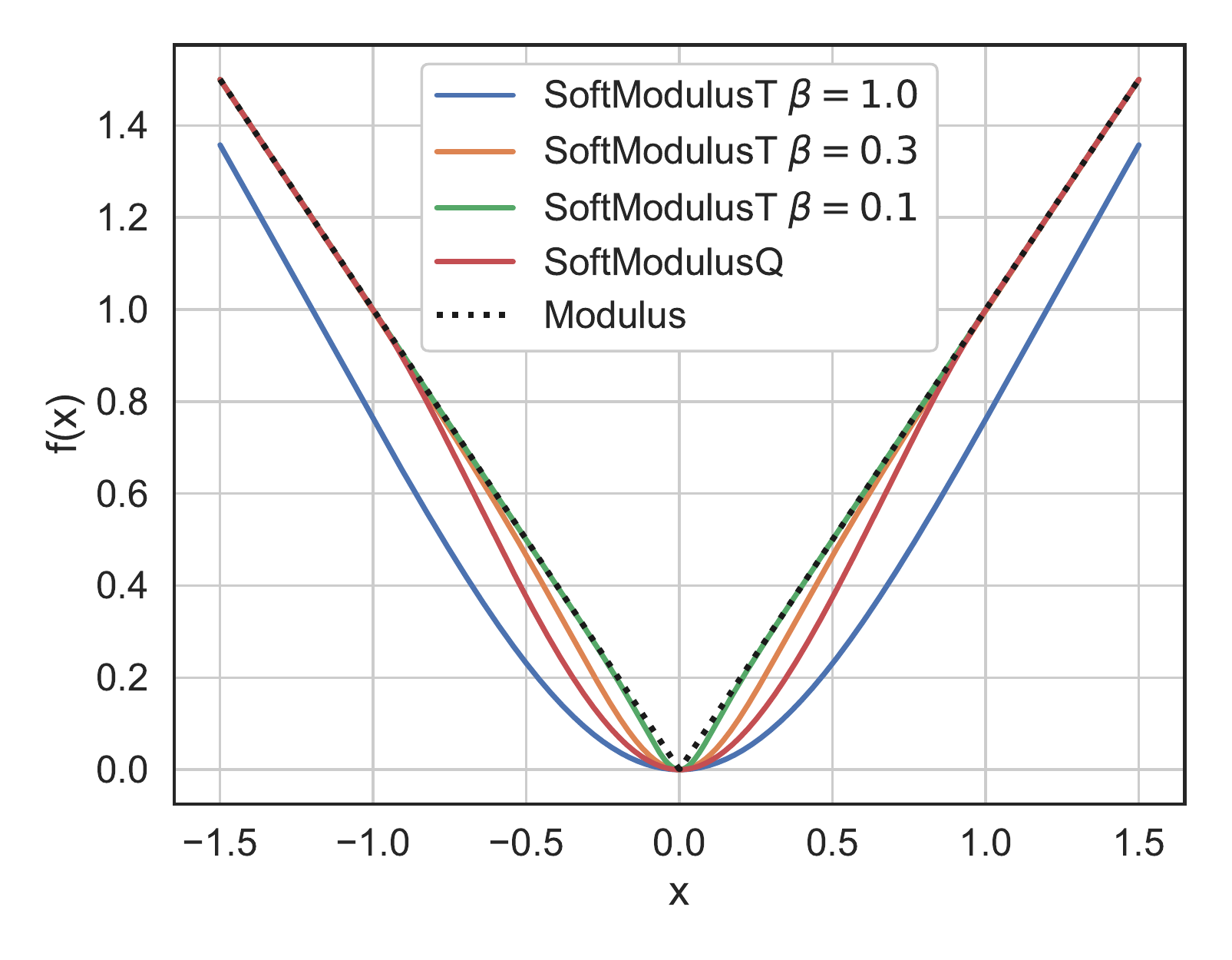}
	\caption{Representation of the two \textit{SoftModulus} activation functions compared with the original \textit{modulus}: \textit{SoftModulusQ} and \textit{SoftModulusT}.}
	\label{fig:activationssmooth}
\end{figure}

The \textit{modulus} function is not differentiable when $x=0$. To study if this property harms the performance of the models in any way, two alternative smooth approximations of the \textit{modulus} function have been tested as an additional experiment. See figure \ref{fig:activationssmooth} for a visual representation. The two approximations are defined below.
\begin{itemize}
	\item Quadratic approximation: referred subsequently as \textit{SoftModulusQ} and defined in equation \ref{eq:quad-approx} (full derivation in the appendix).
	\begin{equation}\label{eq:quad-approx}
	f(x)= \left\{ \begin{array}{lcc}
	x^2 \cdot (2-|x|) &  \text{if} & |x| \leq 1 \\
	|x| &   \text{if}  & |x| > 1
	\end{array}
	\right.
	\end{equation}
	\item Hyperbolic tangent approximation: referred subsequently as \textit{SoftModulusT} and defined in equation \ref{eq:tanh-approx}. Starting from the modulus function $f(x)=sgn(x)\cdot x$, we approximate the $sgn$ function as follows $sgn(x) \approx \tanh(x/\beta)$, where $\beta \in [0, 1]$ is a tunable hyperparameter that controls how acute the ``V'' transition is. The smaller the $\beta$, the closest the approximation is to the \textit{modulus}. We used a value of $\beta=0.01$ in all the models we trained.
	\begin{equation} \label{eq:tanh-approx}
	f(x) = x \cdot \tanh(x/\beta)
	\end{equation}
\end{itemize}

\subsection{Benchmark activation functions}
The following activation functions have been used as a benchmark to compare the performance of the proposed one.

\begin{itemize}
	\item \textit{Tanh}: the hyperbolic tangent has been one of the most popular choices (together with the \textit{sigmoid}), before \textit{ReLU} was proposed \citep{lecun2012}. Among many other desirable properties, its derivative is very simple: $(\tanh x)'=1-\tanh^2 x$. This property was very beneficial specially before automatic differentiation tools appeared. Besides, similar to our proposed activation function, the derivative can be easily calculated during forward pass and cached for the backward pass. 
	\item \textit{ReLU}: this activation function was published in 2010 as an alternative to train \textit{Restricted Boltzman Machines}. It is defined as $f(x) = \max(0,x)$. It allowed training deeper neural networks by solving the vanishing gradient problems typically happening with saturating activation functions. As it can be seen in the formula, ReLU outputs zero if the input is negative. This brings sparsity to the non-linear representation, at the cost of potentially finding optimization problems due to the fact that the derivative when $x<0$ is zero (dying neurons, for instance).
	\item \textit{Leaky-ReLU}: the \textit{Leaky-ReLU} attempted to solve the problem known as \textit{dying neurons} in the \textit{ReLUs} by adding a small linear term in the negative side of $x$ (see equation \ref{eq:leaky-relu}, where $\beta>1$ is a hyperparameter to be tuned).
	\begin{equation} \label{eq:leaky-relu}
		f(x) = \max(x/\beta, x)
	\end{equation}
	\item \textit{ELU}: this is a smooth version of \textit{ReLU} that approaches asymptotically to -1 as $x\rightarrow-\infty$. Unlike ReLUs, it is differentiable everywhere and can produce negative values. This activation function is formally defined in equation \ref{eq:elu}, where $\beta$ is a tunable hyperparameter.
	\begin{equation} \label{eq:elu}
	f(x)= \left\{ \begin{array}{lcc}
		x &   \text{if}  & x > 0 \\
 \beta(e^x - 1) &  \text{if} & x \leq 0 
	\end{array}
	\right.
	\end{equation}
	
	\item \textit{Swish}: this is one of the few non-monotonic activation functions formally published. It gained its popularity because it showed promising results in multiple applications. It is defined as $f(x) = x \cdot \text{sigmoid}(\beta x)$, where $\beta$ is a hyperparameter to be tuned.
	\item \textit{Mish}: this is a smooth, non-monotonic and self-regularized activation function, similar to swish, showing superior performance in different benchmarks. It is defined as $f(x) = x \cdot \tanh (\text{softplus} (x))$, where $\text{softplus} (x) = \log(1+e^x)$ \citep{dugas2001}.
	\item \textit{PFLU}: the Power Function Linear Unit (PFLU) is a non-monotonic activation function published on 2020 that showed good performance in convolutional architectures. It is defined as follows: $f(x) = x \cdot \frac{1}{2} \left( 1 + \frac{x}{\sqrt{1+x^2}} \right)$.
	
\end{itemize}

\section{Experiments and results} \label{sec:experiments}
\subsection{Setup}

\begin{table}[h] \small
	\caption{Deep learning architectures used to experiment with different activations. In the dense layers row, the output size has been represented as $C$. The Fully connected architecture (\textit{FC}) is a multilayer perceptron with 2 hidden layers + the output layer. The \textit{Conv2} and \textit{Conv6} architectures are shallow variants of \textit{VGG} described here: \citep{simonyan2015}. The last pooling layer of the \textit{VGG-16} original architecture has been trimmed in order to allow this model to work with smaller image sizes.}
	\setcellgapes{3pt}\makegapedcells
	\begin{tabular}[t]{lllll}
		\toprule
		Network              & Fully Connected & Conv2       & Conv6                                                & VGG-16                                                                                  \\ \midrule
		\makecell[l]{Conv layers\\\\\\\\\\} &                & \makecell[l]{64,64,pool\\\\\\\\\\}  & \makecell[l]{64,64,pool\\128,128,pool\\256,256,pool\\\\\\} & \makecell[l]{64,64,pool\\128,128,pool\\256,256,256,pool\\512,512,512,pool\\512,512,512} \\
		Dense Layers         & 256,256,$C$     & 256,256,$C$ & 256,256,$C$                                          & 4096,4096,$C$                                                                           \\
		Filter sizes &  & 3x3 & 3x3 & 3x3 \\
		Pooling type & & max & max & max \\
		\# Parameters        & 269k-878k       & 3.3M-4.3M   & 1.8M-2.3M                                            & 33.6M-40.3M                                                                             \\ 			        
		Size                 & 3.1-11MB & 38-50MB & 21-27MB & 385-462MB \\ \bottomrule
	\end{tabular}
	\label{tab:architectures}
	\vspace{20pt}
\end{table}

For the following experiments we have used \textit{CIFAR10}, \textit{CIFAR100} \citep{krizhevsky09} and \textit{MNIST} \citep{lecun2010} datasets. \textit{CIFAR10} and \textit{CIFAR100} images are labeled into 10 and 100 classes, respectively. They are of size 32x32 and full color. In both cases, the data sets contain 50,000 images for training and 10,000 images for testing purposes. \textit{MNIST} images belong to one of 10 classes and are in grey scale and 28x28 size. \textit{MNIST} comes with 50,000 images for training and 10,000 images for testing purposes.

The images of the datasets have been normalized so that the minimum and the maximum values are -1 and 1. We have kept untouched the default train/test split provided by \textit{Pytorch} \citep{Paszke2019}, in order to facilitate future potential reproducibility and benchmarking efforts. 

Four different architectures have been used to test the performance of the \textit{modulus} activation function against the other nonlinearities. These architectures consist of a multilayer perceptron (named \textit{fully connected}), two convolutional architectures with 2 and 6 convolutional layers (named \textit{conv2} and \textit{conv6} respectively) and the \textit{VGG-16} network \cite{simonyan2015}. They are described in more detail in Table \ref{tab:architectures}. The architectures choice has been inspired on the \textit{Lottery Ticket Hypothesis} paper \citep{frankleC19}, however no pruning methods have been applied in this study.

\definecolor{First}{HTML}{A2BEFF}
\definecolor{Second}{HTML}{A8FFD4}
\definecolor{Third}{HTML}{FFEDB8}
\definecolor{Fourth}{HTML}{FFB8C8}

\begin{table}[h!] \footnotesize  \setlength{\tabcolsep}{3pt}
	\caption{Test accuracy for all the datasets and activation functions tested (rows), and for all the models (columns). The results are expressed as mean $\pm$ standard deviation across the 30 random initializations. To facilitate the reading of the table, for each dataset-model combination, we have colored the results as follows: the best result (highest accuracy) has been colored in blue, the second in green, the third in yellow and the fourth in red (\colorbox{First}{first} $>$ \colorbox{Second}{second} $>$ \colorbox{Third}{third} $>$ \colorbox{Fourth}{fourth}). Additionally, we marked in bold those cases where any of the proposed activation functions achieved significantly higher accuracy than the benchmarks, with a significance level of $\alpha=0.05$, for which we used a \textit{Wilcoxon one-sided Rank Sum}  test.}
	\centering
	\begin{tabular}{rrcccc}
		\toprule
		 Dataset &   Activation &                     FC                      &                    Conv2                    &                    Conv6                    &                    VGG16                    \\ \midrule
		 CIFAR10 &         ReLU &              $54.65 \pm 0.22$               &              $71.40 \pm 0.26$               &              $77.09 \pm 1.21$               &     $\cellcolor{Fourth}83.66 \pm 0.41$      \\
		         &    LeakyReLU &              $54.71 \pm 0.24$               &      $\cellcolor{Third}71.65 \pm 0.32$      &              $77.38 \pm 1.13$               &      $\cellcolor{Third}83.98 \pm 0.34$      \\
		         &         Tanh &              $49.95 \pm 0.23$               &              $67.46 \pm 0.34$               &              $77.67 \pm 0.24$               &              $79.69 \pm 0.26$               \\
		         &        Swish &      $\cellcolor{Third}55.39 \pm 0.29$      &              $69.09 \pm 0.27$               &              $71.66 \pm 0.62$               &              $80.77 \pm 0.50$               \\
		         &          ELU &     $\cellcolor{Fourth}55.16 \pm 0.21$      &              $69.34 \pm 0.34$               &              $78.98 \pm 0.33$               &              $81.21 \pm 0.37$               \\
		         &         PFLU &     $\cellcolor{Second}55.43 \pm 0.30$      &              $70.34 \pm 0.33$               &     $\cellcolor{Fourth}80.77 \pm 0.40$      &              $81.58 \pm 0.38$               \\
		         &         Mish &      $\cellcolor{First}55.44 \pm 0.22$      &              $69.66 \pm 0.35$               &              $75.66 \pm 0.53$               &              $80.98 \pm 0.64$               \\
		         &      Modulus &              $53.97 \pm 0.24$               & $\mathbf{\cellcolor{Second}73.93 \pm 0.42}$ & $\mathbf{\cellcolor{Second}84.22 \pm 0.29}$ & $\mathbf{\cellcolor{Second}84.86 \pm 0.32}$ \\
		         & SoftModulusQ &              $54.07 \pm 0.29$               &     $\cellcolor{Fourth}71.49 \pm 0.37$      & $\mathbf{\cellcolor{Third}81.01 \pm 1.27}$  &              $10.00 \pm 0.00$               \\
		         & SoftModulusT &              $54.04 \pm 0.24$               & $\mathbf{\cellcolor{First}73.95 \pm 0.40}$  & $\mathbf{\cellcolor{First}84.36 \pm 0.28}$  & $\mathbf{\cellcolor{First}85.34 \pm 0.36}$  \\ \midrule
		CIFAR100 &         ReLU &              $27.33 \pm 0.25$               &              $36.72 \pm 0.31$               &              $36.35 \pm 0.89$               &              $44.61 \pm 1.11$               \\
		         &    LeakyReLU &              $27.24 \pm 0.22$               &              $37.03 \pm 0.40$               &              $37.15 \pm 0.77$               &              $45.19 \pm 1.47$               \\
		         &         Tanh &              $23.63 \pm 0.20$               &              $35.29 \pm 0.47$               &              $42.15 \pm 0.49$               &              $44.14 \pm 0.37$               \\
		         &        Swish &     $\cellcolor{Fourth}27.59 \pm 0.25$      &              $35.20 \pm 0.34$               &              $35.75 \pm 0.41$               &              $46.02 \pm 1.10$               \\
		         &          ELU &      $\cellcolor{First}27.92 \pm 0.26$      &              $35.68 \pm 0.31$               &              $40.74 \pm 0.48$               &     $\cellcolor{Fourth}47.63 \pm 0.71$      \\
		         &         PFLU &     $\cellcolor{Second}27.73 \pm 0.21$      &      $\cellcolor{Third}37.51 \pm 0.42$      &     $\cellcolor{Fourth}42.25 \pm 0.45$      &      $\cellcolor{Third}48.22 \pm 0.63$      \\
		         &         Mish &      $\cellcolor{Third}27.68 \pm 0.25$      &              $36.04 \pm 0.41$               &              $37.63 \pm 0.75$               &      $\cellcolor{First}48.69 \pm 0.69$      \\
		         &      Modulus &              $26.29 \pm 0.26$               & $\mathbf{\cellcolor{Second}38.66 \pm 0.56}$ & $\mathbf{\cellcolor{First}48.73 \pm 0.62}$  &              $45.83 \pm 0.80$               \\
		         & SoftModulusQ &              $26.23 \pm 0.25$               &     $\cellcolor{Fourth}37.48 \pm 0.44$      & $\mathbf{\cellcolor{Third}48.16 \pm 1.97}$  &               $1.00 \pm 0.00$               \\
		         & SoftModulusT &              $26.32 \pm 0.24$               & $\mathbf{\cellcolor{First}38.69 \pm 0.56}$  & $\mathbf{\cellcolor{Second}48.63 \pm 0.83}$ &     $\cellcolor{Second}48.47 \pm 0.68$      \\ \midrule
		   MNIST &         ReLU &              $98.35 \pm 0.07$               &     $\cellcolor{Fourth}99.27 \pm 0.04$      &              $99.53 \pm 0.03$               &     $\cellcolor{Fourth}99.58 \pm 0.04$      \\
		         &    LeakyReLU &              $98.37 \pm 0.06$               &              $99.27 \pm 0.04$               &              $99.53 \pm 0.03$               &      $\cellcolor{Third}99.58 \pm 0.03$      \\
		         &         Tanh &              $98.34 \pm 0.07$               &              $99.06 \pm 0.05$               &              $99.48 \pm 0.04$               &              $99.48 \pm 0.04$               \\
		         &        Swish &              $98.36 \pm 0.05$               &              $99.24 \pm 0.04$               &              $99.52 \pm 0.03$               &              $99.53 \pm 0.03$               \\
		         &          ELU &              $98.31 \pm 0.04$               &              $99.16 \pm 0.04$               &              $99.54 \pm 0.03$               &              $99.54 \pm 0.03$               \\
		         &         PFLU &     $\cellcolor{Fourth}98.42 \pm 0.05$      &              $99.21 \pm 0.04$               &     $\cellcolor{Fourth}99.56 \pm 0.03$      &              $99.57 \pm 0.03$               \\
		         &         Mish &              $98.41 \pm 0.05$               &              $99.23 \pm 0.04$               &              $99.56 \pm 0.03$               &              $99.57 \pm 0.04$               \\
		         &      Modulus & $\mathbf{\cellcolor{Third}98.47 \pm 0.07}$  & $\mathbf{\cellcolor{Second}99.38 \pm 0.04}$ & $\mathbf{\cellcolor{Third}99.60 \pm 0.03}$  & $\mathbf{\cellcolor{First}99.63 \pm 0.04}$  \\
		         & SoftModulusQ & $\mathbf{\cellcolor{First}98.51 \pm 0.06}$  & $\mathbf{\cellcolor{Third}99.37 \pm 0.03}$  & $\mathbf{\cellcolor{First}99.62 \pm 0.03}$  &              $11.35 \pm 0.00$               \\
		         & SoftModulusT & $\mathbf{\cellcolor{Second}98.47 \pm 0.06}$ & $\mathbf{\cellcolor{First}99.39 \pm 0.04}$  & $\mathbf{\cellcolor{Second}99.61 \pm 0.03}$ & $\mathbf{\cellcolor{Second}99.62 \pm 0.03}$ \\ \bottomrule
	\end{tabular}	
	
	\label{tab:results}
\end{table}

\begin{figure}[h!]
	\centering
	\includegraphics[width=1.0\linewidth]{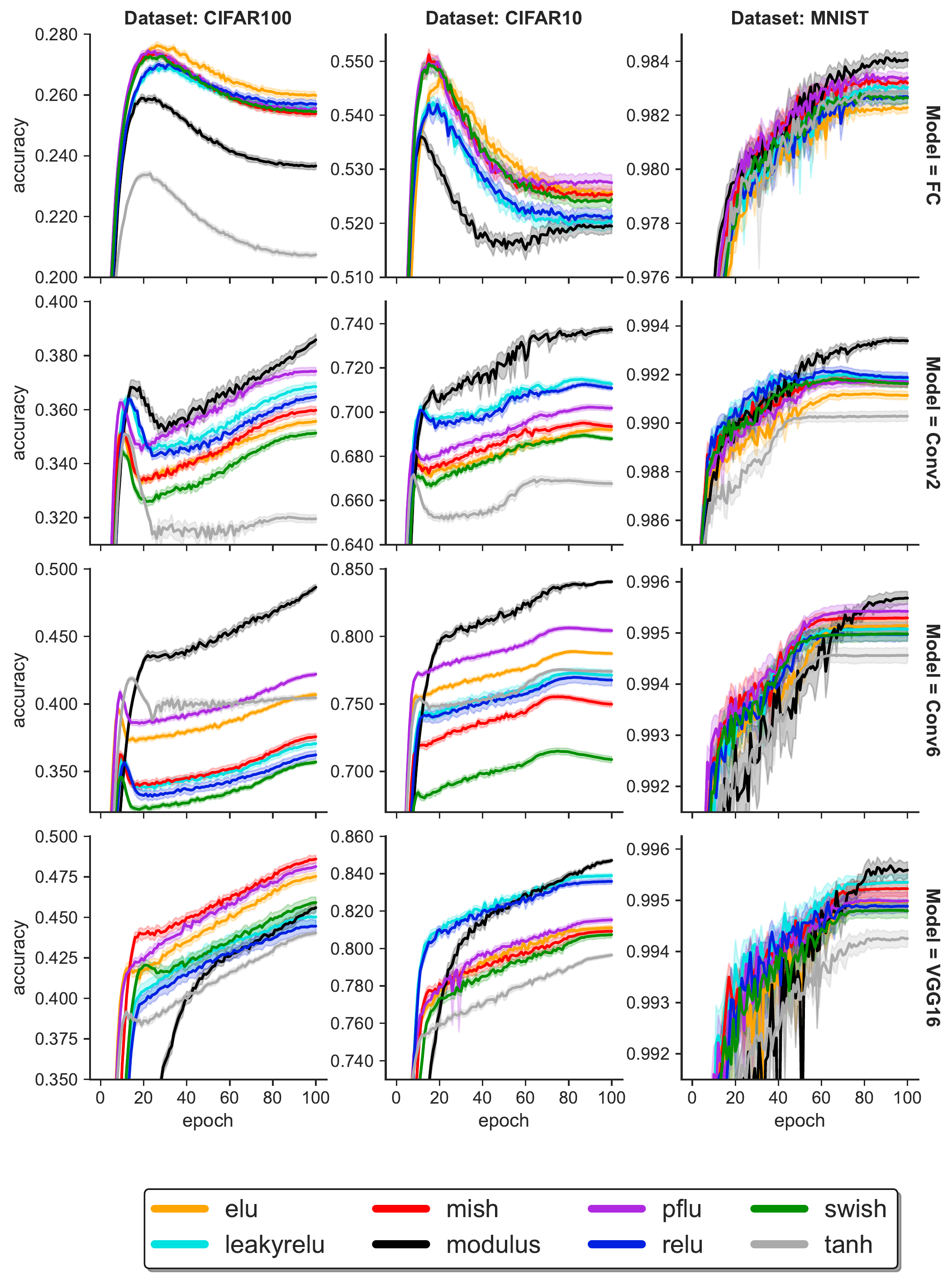}
	\caption{Learning curves of all the trained models. Each line represents the average test accuracy of 30 runs of a model. The shade behind to each line shows the 95\% confidence interval around the mean. Notice that the scales in the y-axis are not shared to allow for better zoom in each figure, comparing the performance across models is not the objective of this study. Details like the shades are better viewed in a screen.}
	\label{fig:training_curves}
\end{figure}

We have followed the same experimental setup across models, activations and datasets. All the networks have been trained for 100 epochs, and the best accuracy of each run has been reported. Each run has been repeated 30 times with different random weight initializations. No \textit{dropout} \citep{Srivastava2014} nor \textit{batch normalization} \citep{Ioffe2015} have been used. We used \textit{Adam} \citep{Kingma14} as optimizer, with a learning rate of $10^{-4}$, a \textit{gradual warmup} \citep{gotmare2018} during the first 5 epochs starting from $10^{-5}$, and a \textit{cosine annealing} \citep{loshchilov2017} with a target learning rate of $10^{-6}$ in the last epoch. We used \textit{Python} 3.7.3, \textit{Pytorch} 1.7.1 and  \textit{TorchVision} 0.8.2, and all the models have been trained in a single \textit{Nvidia 2080ti} graphics card. The code used can be found in the url of the footnote\footnote{\url{https://github.com/ivallesp/abs}}.

\subsection{Classification performance}
Table \ref{tab:results} summarizes the results of the classifiers for \textit{CIFAR10}, \textit{CIFAR100} and \textit{MNIST} datasets. As we can see from the tables, the \textit{modulus} activation function outperforms significantly the benchmark activations in 9 out of 12 experiments (4 architectures x 3 datasets). In 4 of these cases, the accuracy improvement is $3\%$ or higher, in relative terms (CIFAR10-Conv2, CIFAR10-Conv6, CIFAR100-Conv2 and CIFAR100-Conv6 with p-value  $<10^{-6}$ in all the cases). If we consider the soft approximations of the modulus activation function, we see that the \textit{SoftModulusT} significantly outperforms the benchmark in 10 out of 12 experiments (with p-value of $6.55 \cdot 10^{-6}$ for MNIST-Conv6, $1.29 \cdot 10^{-2}$ for MNIST-VGG16, $3.99 \cdot 10^{-4}$ for CIFAR100-VGG16 and $<10^{-6}$ for the rest). All the p-values reported in this study have been obtained using a \textit{Wilcoxon one-sided Rank Sum} test comparing the modulus results against the best of the benchmark activation functions in each case. 

\begin{figure}[h!]
	\centering
	\includegraphics[width=1.0\linewidth]{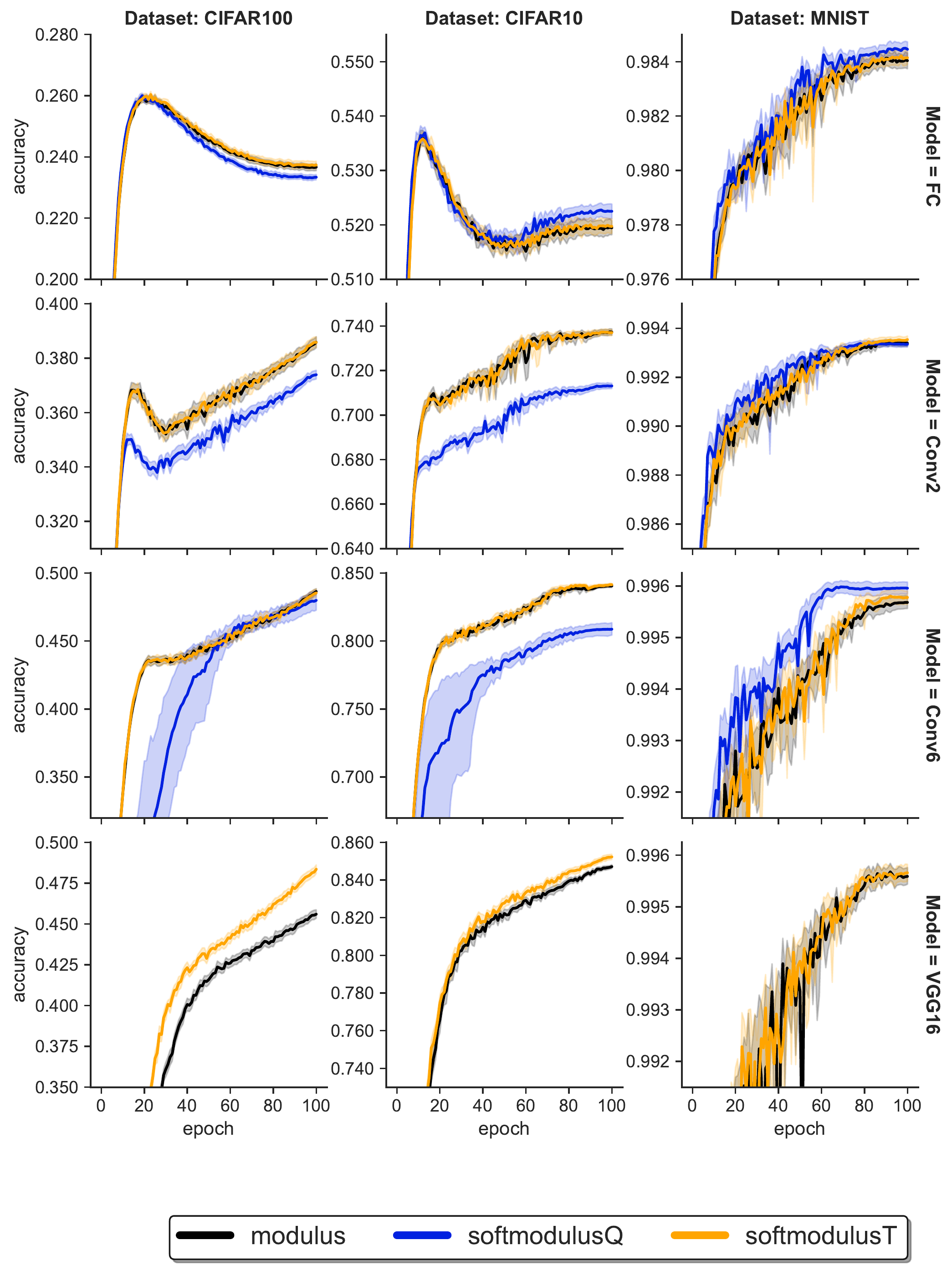}
	\caption{Learning curves for smooth approximations compared with the \textit{modulus}. Each line represents the average test accuracy of 30 runs of a model. The shade behind to each line shows the 95\% confidence interval around the mean. Notice that the scales in the y-scales are not shared to allow for better zoom in each figure, comparing the performance across models is not the objective of this study. Details like the shades are better viewed in a screen.}
	\label{fig:training_curves_smooth}
\end{figure}

Figure \ref{fig:training_curves} shows the test accuracy at every epoch for the 12 experiments (the curves of the smooth approximations have been included in figure \ref{fig:training_curves_smooth} to compare them with the \textit{modulus}). As it can be noticed in several cases (e.g. \textit{CIFAR10-Conv2}, \textit{CIFAR10-Conv6}, \textit{CIFAR100-Conv6}), the accuracy of the networks with \textit{modulus} activation function keeps increasing at the 100th epoch while the benchmarks have already stabilized. That observation suggests that, if trained for longer, probably the accuracy would increase even more.

The results of the \textit{SoftModulus} approximations compared with the \textit{modulus} activation function are summarized in table \ref{tab:results_smooth}. Additionally, figure \ref{fig:training_curves_smooth} shows the test accuracy at every epoch for the same activations. From these results, we see that although the \textit{SoftModulusQ} approximation achieves significantly better results than the original modulus in several cases (\textit{FC} model over \textit{CIFAR10} and \textit{MNIST} dataset, and \textit{Conv6} over \textit{MNIST} dataset), it seems to be much more unstable: on VGG16, the gradients of the models with \textit{SoftModulusQ} activations vanish at the beginning of the training process due to the fact that the weights are initialized very close to zero and the gradients are too small, causing numerical problems. This problem may be solved tweaking the initialiation. However, the \textit{SoftModulusT} approximation seems to be on-par with the original \textit{modulus} in the majority of cases, while it perform significantly better for deep architectures like \textit{VGG16}. We hypothesize that the difference between the two approximations is due to the width of the zero gradient region around $x=0$: wider regions lead to more training problems (see figure \ref{fig:activationssmooth}). The $\beta$ parameter in the hyperbolic tangent approximation allows for easily adjust this zero-gradient region. We informally tested different values of beta concluding that for high values of $\beta$ (e.g. $\beta=1.0$) the model struggles to train due to numerical precision problems (small values of weights lead to tiny gradients). A value of $\beta=0.01$ seems to work well for all the tested experiments. Finally, we see that the \textit{SoftModulusT} achieves superior results than the original \textit{modulus} when used in the \textit{VGG16} architecture.

\begin{table}[h!] \footnotesize  \setlength{\tabcolsep}{3pt}
	\caption{Accuracy comparison for the soft approximations of the \textit{modulus} function, compared with the original \textit{modulus} definition. The results are expressed as mean $\pm$ standard deviation across the 30 random initializations. In each column we highlighted the model with higher accuracy in bold. We added a star to those cases where a \textit{SoftModulus} activation function achieved significantly higher results than the \textit{modulus}, with a significance level of $\alpha=0.05$.}
	\centering
	\begin{tabular}{rrcccc}
		\toprule
		 Dataset &   Activation &            FC             &           Conv2           &           Conv6           &           VGG16           \\ \midrule
		 CIFAR10 &      Modulus &     $53.97 \pm 0.24$      &     $73.93 \pm 0.42$      &     $84.22 \pm 0.29$      &     $84.86 \pm 0.32$      \\
		         & SoftModulusQ & $\mathbf{54.07 \pm 0.29}*$ &     $71.49 \pm 0.37$      &     $81.01 \pm 1.27$      &     $10.00 \pm 0.00$      \\
		         & SoftModulusT &     $54.04 \pm 0.24$      & $\mathbf{73.95 \pm 0.40}$ & $\mathbf{84.36 \pm 0.28}$ & $\mathbf{85.34 \pm 0.36}*$ \\ \midrule
		         
		CIFAR100 &      Modulus &     $26.29 \pm 0.26$      &     $38.66 \pm 0.56$      & $\mathbf{48.73 \pm 0.62}$ &     $45.83 \pm 0.80$      \\
		         & SoftModulusQ &     $26.23 \pm 0.25$      &     $37.48 \pm 0.44$      &     $48.16 \pm 1.97$      &      $1.00 \pm 0.00$      \\
		         & SoftModulusT & $\mathbf{26.32 \pm 0.24}$ & $\mathbf{38.69 \pm 0.56}$ &     $48.63 \pm 0.83$      & $\mathbf{48.47 \pm 0.68}*$ \\ \midrule
		         
		   MNIST &      Modulus &     $98.47 \pm 0.07$      &     $99.38 \pm 0.04$      &     $99.60 \pm 0.03$      & $\mathbf{99.63 \pm 0.04}$ \\
		         & SoftModulusQ & $\mathbf{98.51 \pm 0.06}*$ &     $99.37 \pm 0.03$      & $\mathbf{99.62 \pm 0.03}*$ &     $11.35 \pm 0.00$      \\
		         & SoftModulusT &     $98.47 \pm 0.06$      & $\mathbf{99.39 \pm 0.04}$ &     $99.61 \pm 0.03$      &     $99.62 \pm 0.03$      \\ \bottomrule
	\end{tabular}
	\label{tab:results_smooth}
\end{table}

\section{Discussion}

The \textit{modulus} activation function has shown promising results in certain contexts, but it is worth discussing potential criticisms of this approach.


One concern may be that the \textit{modulus} activation does not seem to be biologically inspired by the way in which axon hillocks fire in biological neurons. Despite this, the primary goal of an activation function is to introduce non-linearity into the neural networks, and the \textit{modulus} activation is effective in achieving this. Additionally, the \textit{modulus} produces non-zero gradients on the negative side, which can be beneficial to avoid vanishing gradients and dying neurons.

Despite the intuition that non-monotonic activations can lead to neurons which parameters are more difficult to optimize, there are several examples in the literature (as mentioned in the introduction) that show that when using those nonlinearities the models tend to achieve better results. One potential explanation for this is that non-monotonic activations can introduce more complex nonlinearities into the network, allowing it to model complex relationships in the data more easily, potentially improving the performance of the resulting model. Another possible reason is that non-monotonic activation functions may have improved gradient properties, making optimization easier. For instance, certain non (strictly) monotonic activation functions, such as \textit{ReLU} and the \textit{modulus}, have very simple gradients.

Finally, another  criticism of the \textit{modulus} function may be its potential to introduce symmetry in neural networks, which may result in oscillation or instability. While in our experiments we did not observe evidences of this behavior, it may be a contributing factor to the slower convergence of networks with the \textit{modulus} function (see figure \ref{fig:training_curves}). Another potential explanation for this observation is the lack of initialization methods specially designed for the \textit{modulus} activation. It is worth noting that symmetry in neural networks can lead to duplication of ways to approximate the desired function, which translates to multiple equally valid solutions in the loss landscape. This may be advantageous, as it results in more regions of the loss landscape containing feasible solutions. Other potential advantage is that symmetry may improve the generalization ability of the network, as it allows to learn patterns that are invariant under certain transformations (e.g. input sign flip in dense layers without bias terms $f_\theta(X) = f_\theta(-X)$). The symmetry property introduced by the \textit{modulus} can have more advantages, disadvantages or applications. Further research could explore the potential side effects of symmetry in more depth, as well as its impact on the performance and stability of the network in other tasks and applications.

\section{Conclusions} \label{sec:conclusions}
We have shown how the new \textit{modulus} activation function outperforms the benchmark activation functions in 75\% of our experiments. The improvement achieved by the networks with this activation function is often significantly higher ($3\%$ or higher in one third of the experiments we conducted). Additionally, we have proposed a smooth version of the \textit{modulus} activation function which performs better than the first one, at a slightly higher computational cost. 

As a next step, it would be beneficial to evaluate the proposed \textit{modulus} activation function on a wider range of problems beyond image classification. This would provide insight into the generalizability and potential utility of the \textit{modulus} function in a variety of tasks. Further investigation in this direction could help to identify the specific areas in which the \textit{modulus} function is especially effective, and highlight new applications for this non-monotonic activation function.

\section{Acknowledgements}
This work has been partially funded by the Spanish ministry of science and innovation MCIN/AEI projects PID2019-107347RR-C33 and PID2021-127946OB-I00.

\newpage
\bibliography{doc_arxiv}
\newpage

\section*{Appendix: SoftModulusQ derivation}
\begin{figure}[h!]
	\centering
	\includegraphics[width=1\linewidth]{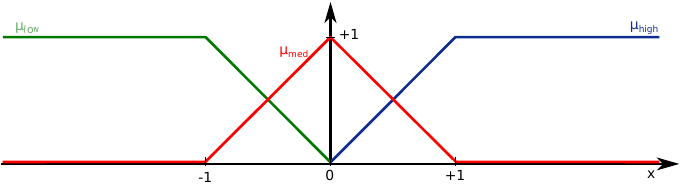}
	\caption{Membership functions used to combine the quadratic and modulus functions.)}
	\label{fig:fuzzy}
\end{figure}

We chose a fuzzy set approach to combine the modulus function with a quadratic function. For that, three membership functions are defined in figure \ref{fig:fuzzy} and equations \ref{eq:mulow}, \ref{eq:mumid}, \ref{eq:muhigh}. 

\begin{equation}
	\label{eq:mulow}
	\mu^x_{\text{low}}= \left\{ \begin{array}{lcc}
		1 &   \text{if}  & x < -1 \\
		-x & \text{if}  & -1 \leq x < 0 \\
		0 &  \text{if} & x \geq 0 
		\end{array}
		\right.
\end{equation}

\begin{equation}
	\label{eq:mumid}
	\mu^x_{\text{med}}= \left\{ \begin{array}{lcc}
		0 &   \text{if}  & x < -1 \\
		x+1 & \text{if}  & -1 \leq x < 0 \\
		1-x & \text{if}  & 0 \leq x < 1 \\
		0 &  \text{if} & x \geq 1
		\end{array}
		\right.
\end{equation}

\begin{equation}
	\label{eq:muhigh}
	\mu^x_{\text{high}}= \left\{ \begin{array}{lcc}
	0 &   \text{if}  & x < 0 \\
	x & \text{if}  & 0 \leq x < 1 \\
	1 &  \text{if} & x \geq 1 
	\end{array}
	\right.
\end{equation}

Then we define $f^x_{\text{low}}=-x$, $f^x_{\text{med}}=x^2$ and $f^x_{\text{high}}=x$. If we combine these functions with the membership functions we get the following.

\begin{itemize}
	\item For $x < -1$: we get $\hat{f}(x)=\frac{f^x_{\text{low}} \cdot \mu^x_{\text{low}}}{\mu^x_{\text{low}}}=-x$

	\item For $-1\leq x < 0$: we get $\hat{f}(x)=\frac{f^x_{\text{low}} \cdot \mu^x_{\text{low}} + f^x_{\text{med}} \cdot \mu^x_{\text{med}}}{\mu^x_{\text{low}} + \mu^x_{\text{med}}}=\frac{(x+1)x^2+(-x)(-x)}{1+x-x}=x^3 + 2x^2$

	\item For $0\leq x < 1$: we get $\hat{f}(x)=\frac{f^x_{\text{med}} \cdot \mu^x_{\text{med}} + f^x_{\text{high}} \cdot \mu^x_{\text{high}}}{\mu^x_{\text{med}} + \mu^x_{\text{high}}}=\frac{(1-x)x^2+x\cdot x}{1-x+x}=-x^3 + 2x^2$

	\item For $x \geq 1$: we get $\hat{f}(x)=\frac{f^x_{\text{high}} \cdot \mu^x_{\text{high}}}{\mu^x_{\text{high}}}=x$
\end{itemize}

Combining and simplifying all the above expressions we get the final \textit{SoftModulusQ} activation function.

\begin{equation}
f(x)= \left\{ \begin{array}{lcc}
x^2 \cdot (2-|x|) &  \text{if} & |x| \leq 1 \\
|x| &   \text{if}  & |x| > 1
\end{array}
\right.
\end{equation}

\end{document}